\documentclass{article}

\usepackage{times}
\usepackage{amssymb}

\usepackage{graphicx} 
\usepackage{subfigure} 

\usepackage{natbib}

\usepackage{algorithm}
\usepackage{algorithmic}

\usepackage[accepted]{icml2014} 

\icmltitlerunning{Deep Belief Nets for Topic Modeling}

\begin{document} 

\twocolumn[
\icmltitle{Deep Belief Nets for Topic Modeling \\ 
           Workshop on Knowledge-Powered Deep Learning for Text Mining (KPDLTM-2014)}

\icmlauthor{Lars Maaloe}{larsma@dtu.dk}
\icmladdress{DTU Compute, Technical University of Denmark (DTU) B322, DK-2800 Lyngby}
\icmlauthor{Morten Arngren}{moa@issuu.com}
\icmladdress{Issuu, Gasv\ae rksvej 16, 3., DK-1656 Copenhagen}
\icmlauthor{Ole Winther}{owi@imm.dtu.dk}
\icmladdress{Cognitive Systems, DTU Informatics, Technical University of Denmark (DTU) B321, DK-2800 Lyngby}

\icmlkeywords{deep belief nets, topic models, neural networks, restricted boltzmann machine, replicated softmax, deep autoencoders, dimensionality reduction, pattern recognition}

\vskip 0.3in
]

\begin{abstract} 
Applying traditional collaborative filtering to digital publishing is challenging because user data is very sparse due to the high volume of documents relative to the number of users. Content based approaches, on the other hand, is attractive because textual content is often very informative. In this paper we describe large-scale content based collaborative filtering for digital publishing. To solve the digital publishing recommender problem we compare two approaches: latent Dirichlet allocation (LDA) and deep belief nets (DBN) that both find low-dimensional latent representations for documents. Efficient retrieval can be carried out in the latent representation. We work both on public benchmarks and digital media content provided by Issuu, an online publishing platform. This article also comes with a newly developed deep belief nets toolbox for topic modeling tailored towards performance evaluation of the DBN model and comparisons to the LDA model.

\end{abstract} 

\section{Introduction}\label{sec:introduction}
This article concerns the comparison of deep belief nets (DBN) and latent Dirichlet allocation (LDA) for finding a low-dimensional latent representation of documents. DBN and LDA are both generative bag-of-words models and represent conceptual meanings of documents. Similar documents to a query document are retrieved from the low-dimensional output space through a distance measurement. A deep belief net toolbox (DBNT)\footnote{Refer to Github.com \textit{Deep Belief Nets for Topic Modeling}.} has been developed to implement the DBN and evaluate comparisons. The advantage of the DBN is that it has the ability of a highly non-linear dimensionality reduction, due to its \textit{deep} architecture \cite{Hinton2006reducingthe}. A very low-dimensional representation in output space results in a fast retrieval of similar documents to a query document. The LDA model is a mixture model seeking to find the posterior distribution between its visible and hidden variables \cite{Blei2003latentdirichlet}. The number of topics $K$ must be given for the LDA model defining the dimensionality of the Dirichlet-distributed output space. The latent representation of a document is the probability for the document to be in each topic, comprising of a vector of size $K$. To run simulations on the LDA model, we have used the Gensim package for Python\footnote{http://radimrehurek.com/gensim/models/ldamodel.html.} \cite{rehurek_lrec}. The article is conducted in collaboration with Issuu\footnote{http://issuu.com}, a digital publishing platform delivering reading experiences of magazines, books, catalogs and newspapers.



\section{Deep Belief Nets}
The DBN is a direct acyclic graph except from the top two layers that form an undirected bipartite graph. The top two layers is what gives the DBN the ability to unroll into a deep autoencoder (DA) and perform reconstructions of the input data \cite{Bengio-2009}. The DBN consist of a visible layer, output layer and a number of hidden layers. The training process of the DBN is defined by two steps: \textit{pre-training} and \textit{fine-tuning}. In pre-training the layers of the DBN are separated pairwise to form \textit{restricted Boltzmann machines} (RBM). Each RBM is trained independently, such that the output of the lower RBM is provided as input to the next higher-level RBM and so forth. This way the layers of the DBN are trained as partly independent systems. The goal of the pre-training process is to achieve approximations of the model parameters. A document is modeled by its word count vector. To model the word-count vectors the bottom RBM is a \textit{replicated softmax model} (RSM) \cite{Salakhutdinov2010replicatedsoftmax}. The hidden layers of the RBMs consist of \textit{stochastic binary units}. Training are executed through \textit{Gibbs sampling} using contrastive divergence as the approximation to the gradient \cite{Hinton2002trainingproducts}. The RBMs applies to \textit{batch learning} and the model only performs a single Gibbs step before updating the weights \cite{Hinton2012apractical}. Given a visible input vector $\hat{v} = [v_1,...,v_D]$ the probability of a hidden unit $j$ is given by
\begin{equation}
p(h_j=1 | \hat{v}) = \sigma (a_j + \sum_{i=1}^D v_i W_{ij}),
\end{equation}
where $\sigma$ denotes the logistic sigmoid function, $a_j$ the bias for the hidden unit $j$, $v_i$ the state of visible unit $i$, $W_{ij}$ the weight between visible unit $i$ and hidden unit $j$ and $D$ denotes the number of visible units. Except for the RSM, the visible units are binary, where the probability is given by
\begin{equation}
p(v_i=1 | \hat{h}) = \sigma (b_i + \sum_{j=1}^M h_j W_{ij}),
\end{equation}
where $b_i$ denotes the bias of visible unit $i$ and $M$ the number of hidden units. The RSM assumes a multinomial distribution where the units of the visible layer are softmax units. Having a number of softmax units with identical weights is equivalent to having one multinomial unit sampled the same number of times \cite{Salakhutdinov2010replicatedsoftmax}. The probability of $v_i$ taking on value $n$ is
\begin{equation}
p(v_i = n | \hat{h}) = \frac{e^{b_{i} + \sum_{j=1}^M h_j W_{ij}}}{\sum_{q=1}^D e^{b_{q} + \sum_{j=1}^M h_j W_{qj}}}.
\end{equation}
The RSM consider the number of words in each document by scaling the bias terms of the hidden units with the length of each document. The weights and biases of the RBM are updated by
\begin{equation}
\Delta W = \epsilon (\mathbb{E}_{p_{data}}[\hat{v}\hat{h}^T]-\mathbb{E}_{p_{recon}}[\hat{v}\hat{h}^T]),
\end{equation}
\begin{equation}
\Delta \hat{b} = \epsilon (\mathbb{E}_{p_{data}}[\hat{h}]-\mathbb{E}_{p_{recon}}[\hat{h}]),
\end{equation}
\begin{equation}
\Delta \hat{a} = \epsilon (\mathbb{E}_{p_{data}}[\hat{v}]-\mathbb{E}_{p_{recon}}[\hat{v}]),
\end{equation}
where $\epsilon$ is the learning rate and the distribution denoted $p_{recon}$ defines the reconstruction of the input data $p_{data}$ and is the result of a Gibbs chain running a single Gibbs step. $\mathbb{E}_{p_{data}}[\cdot]$ is the expectation with respect to the joint distribution of the real data $p_{data}(\hat{h},\hat{v}) = p_{data}(\hat{h}|\hat{v})p_{data}(\hat{v})$. $\mathbb{E}_{p_{recon}}[\cdot]$ denotes the expectation with respect to the reconstructions. To optimize the training we add weight decay and momentum to the parameter update \cite{Hinton10discoveringbinary}. The model parameters from pre-training is passed on to the fine-tuning. The network is transformed into a DA, by replicating and mirroring the input and hidden layers and attaching them to the output of the DBN. Backpropagation on unlabeled data can be performed on the DA, by computing a probability of the input data $p(\hat{x})$ instead of computing the probability of a label $\hat{t}$ provided the input data $p(\hat{t}|\hat{x})$. This way it is possible to generate an error estimation by comparing the normalized input data to the output probability. The stochastic binary units of the pre-training is replaced by sigmoid units with deterministic, real-valued probabilities. Since the input data is under a multinomial distribution, \textit{cross-entropy} is applied as the error function. The \textit{conjugate gradient} optimization framework is used to produce new values of the model parameters that will ensure convergence. The DBN can output binary and real output values \cite{Salakhutdinov2009semantichashing}. The binary output values are computed by adding deterministic Gaussian noise to the input of the output layer during fine-tuning. This way the output of the logistic sigmoid function at the output units will be close to 0 or 1 \cite{Salakhutdinov2009semantichashing}. The output values of the trained DBN are compared to a threshold\footnote{A threshold of 0.1 is applied due to a high number of output values lying closer to 0 than 1 \cite{Hinton10discoveringbinary}.} in order to decide the binary value. Distance metrics when using binary output vectors are much faster \cite{Hinton10discoveringbinary}.

\section{Simulations}
We have performed model evaluations on the 20 Newsgroups dataset\footnote{Refer to \cite{Hinton10discoveringbinary} for details.}. A dataset based on the \textit{Wikipedia Corpus} is used to compare the DBN to the LDA model, since it contains labeled data. The \textit{Issuu Corpus} has no labeled test set, so we compare the DBN to labels defined by a human perception of the topic distributions of Issuu's LDA model. The models are evaluated by retrieving a number of similar documents to a query document in the test set and average over all possible queries. This provides a fraction of the number of documents in the test set having similar documents in their proximity in the output space\footnote{Euclidean distance and hamming distance is applied as distance metric on real valued and binary output vectors respectively.}. The number of neighbors evaluated are $1, 3, 7, 15, 31,$ and $63$. The evaluation is denoted the \textit{accuracy measurement}.

The learning rate is set to $\epsilon = 0.01$, momentum $m = 0.9$ and a weight decay $\lambda = 0.0002$. The weights are initialized from a $0$-mean normal distribution with variance $0.01$. The biases are initialized to $0$ and the number of epochs are set to $50$. The pre-training procedure applies to \textit{batch}-learning, where each batch represents $100$ documents. For fine-tuning, larger batches of $1000$ documents are generated. We perform three line searches for the conjugate gradient algorithm and the number of epochs is set to 50. The Gaussian noise for the binary output DBN, is defined as deterministic noise with mean $0$ and variance $16$ \cite{Hinton10discoveringbinary}.

\subsection{Model Evaluation}
\begin{figure}[!ht]
   \centering
   \includegraphics[scale=0.35]{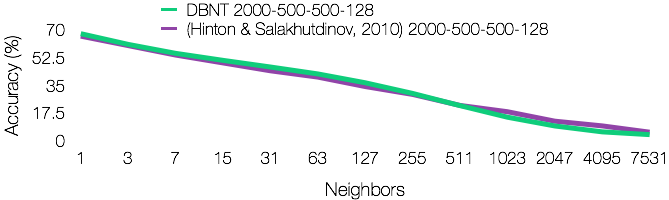}
   \caption{Accuracy measurements of the 2000-500-500-128-DBN with binary output units from \cite{Hinton10discoveringbinary} and a 2000-500-500-128-DBN with binary output units from the DBNT. The models are trained on the 20 Newsgroups dataset. \textbf{NB:} The results from \cite{Hinton10discoveringbinary} are read directly of the graph.}
   \label{fig:sanity}
\end{figure}

\begin{figure}[!ht]
   \centering
   \includegraphics[scale=0.35]{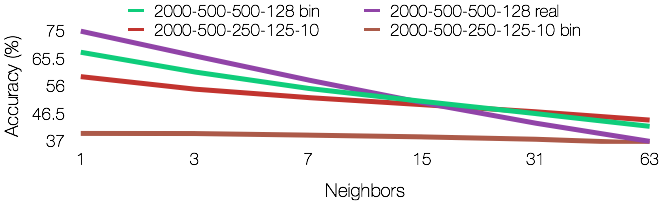}
   \caption{Accuracy measurements of two 2000-500-500-128-DBNs with binary output units and real valued output units trained on the 20 Newsgroups dataset.}
   \label{fig:sanity2}
\end{figure}

\begin{figure}[!ht]
   \centering
   \includegraphics[scale=0.35]{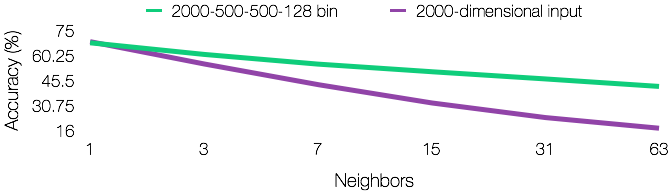}
   \caption{Accuracy measurements of the 2000-500-500-128-DBN with binary output units and the 2000-dimensional input vectors.}
   \label{fig:input_output_comp}
\end{figure}

From Fig. \ref{fig:sanity} the DBNT performs in comparison to the model by Hinton and Salakhutdinov in \cite{Hinton10discoveringbinary}. When comparing the real valued output DBN with a binary output DBN, we have observed that the accuracy measurements are very similar for a higher dimensional output vector (cf. Fig. \ref{fig:sanity2}). For the following simulations we have only considered real valued output vectors though. Fig. \ref{fig:input_output_comp} shows that the DBN manages to find an internal representation of the documents that are better than the high dimensional input vectors.

\subsection{Wikipedia Corpus}
\begin{figure}[!ht]
   \centering
   \includegraphics[scale=0.35]{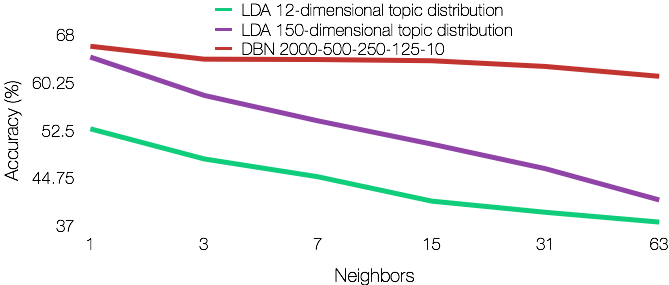}
   \caption{Accuracy measurements of two LDA models and a 2000-500-250-125-10 DBN.}
   \label{fig:dbn_lda}
\end{figure}

We have generated a dataset based on the Wikipedia Corpus. The dataset is denoted \textit{Wikipedia Business} and contain articles from 12 subcategories from the \textit{Business} category. We will use categories with a large pool of articles and a strong connectivity to the remaining categories of the Wikipedia Corpus. The categories are \textit{administration}, \textit{commerce}, \textit{companies}, \textit{finance}, \textit{globalization}, \textit{industry}, \textit{labor}, \textit{management}, \textit{marketing}, \textit{occupations}, \textit{sales} and \textit{sports business}. The Wikipedia Business dataset consists of $32843$ documents split into $22987$ (70\%) training set documents and $9856$ (30\%) test set documents. Wikipedia Business provide an indication on how well the DBN and LDA model captures the granularity of the data within sub-categories of the Wikipedia Corpus. In order to compare the DBN model to the LDA model, we have computed accuracy measurements on a 2000-500-250-125-10-DBN with real numbered linear output units and accuracy measurements on two LDA models, one with $K=12$ topics and another with $K=150$ topics. The accuracy measurement of the 2000-500-250-125-10-DBN is outperforming the two LDA models (cf. Fig. \ref{fig:dbn_lda}). The LDA model with $K=12$ topics perform much worse than the DBN. The LDA model with a $K=150$ topics perform well when evaluating 1 neighbor, but deteriorates quickly throughout the evaluation points. The DBN is the superior model for dimensionality reduction on the Wikipedia Business dataset. Its accuracy measurements are higher and the output is $10$-dimensional compared to the $150$-dimensional topic distribution of the LDA model with the lowest error.

\begin{figure}[!ht]
   \centering
   \includegraphics[scale=0.35]{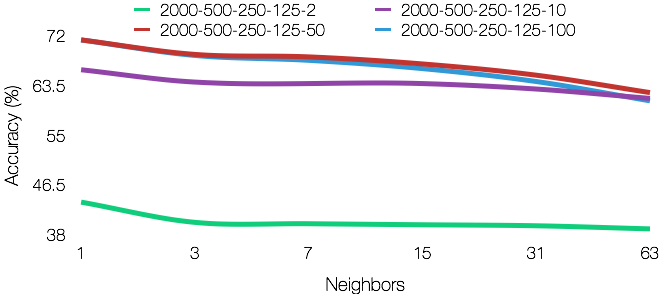}
   \caption{Accuracy measurements on DBNs with different number of output units.}
   \label{fig:output_dbn}
\end{figure}

\begin{figure}[!ht]
   \centering
   \includegraphics[scale=0.30]{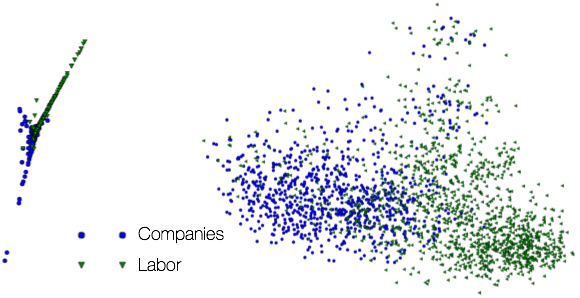}
   \caption{PCA on the 1st and 2nd principal components on the test dataset input vectors and output vectors from a 2000-500-250-125-10-DBN. \textbf{Left:} PCA on the $2000$-dimensional input. \textbf{Right:} PCA on the $10$-dimensional output.}
   \label{fig:wiki_pca}
\end{figure}

We have computed accuracy measurements for: 2000-500-250-125-2-DBN, 2000-500-250-125-10-DBN, 2000-500-250-125-50-DBN and 2000-500-250-125-100-DBN (cf. Fig. \ref{fig:output_dbn}). It is evident that the DBN with a 2-dimensional output scores a much lower accuracy measurement, due to its inability to differentiate between the documents. When increasing the number of output units by modeling the 2000-500-250-125-50-DBN and the 2000-500-250-125-100-DBN, we see that they outperform the original 2000-500-250-125-10-DBN. Even though one DBN has an output vector twice the size of the other, the two evaluations are almost identical, which indicates saturation.
Fig. \ref{fig:wiki_pca} shows how the DBN spreads the data in output space. Since PCA has its limitations it is not possible to visualize more categories unless an approach such as t-SNE is applied \cite{citeulike:3749741}.

\subsection{Issuu Corpus}
\begin{figure}[!ht]
   \centering
   \includegraphics[scale=0.35]{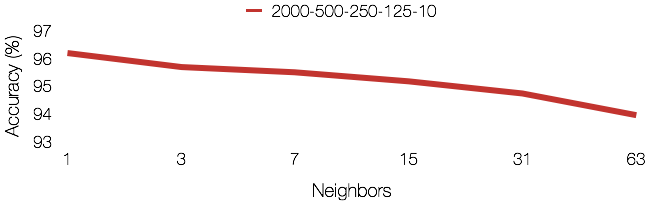}
   \caption{Accuracy measurements of a 2000-500-250-125-10-DBN on the labels defined on the basis of Issuu's LDA model.}
   \label{fig:acc}
\end{figure}

To test the DBN on the Issuu dataset we have extracted a dataset across 5 categories defined from Issuu's LDA model. The documents in the dataset belong to the categories \textit{Business}, \textit{Cars}, \textit{Food \& Cooking}, \textit{Individual \& Team Sports} and \textit{Travel}. The training set contains $13650$ documents and the test set contains $5850$ documents. As mentioned, Issuu has applied labels to the dataset from the results of their LDA model with a $150$-dimensional latent representation. In order to compare the models, we have performed accuracy measurements for the 2000-500-250-125-10-DBN on these labels (cf. Fig. \ref{fig:acc}). From the accuracy measurements it is evident how similar the results of the two models are. The big difference is that the DBN generates a $10$-dimensional latent representation as opposed to the $150$-dimensional latent representation of the LDA model.

\begin{figure}[!ht]
   \centering
   \includegraphics[scale=0.20]{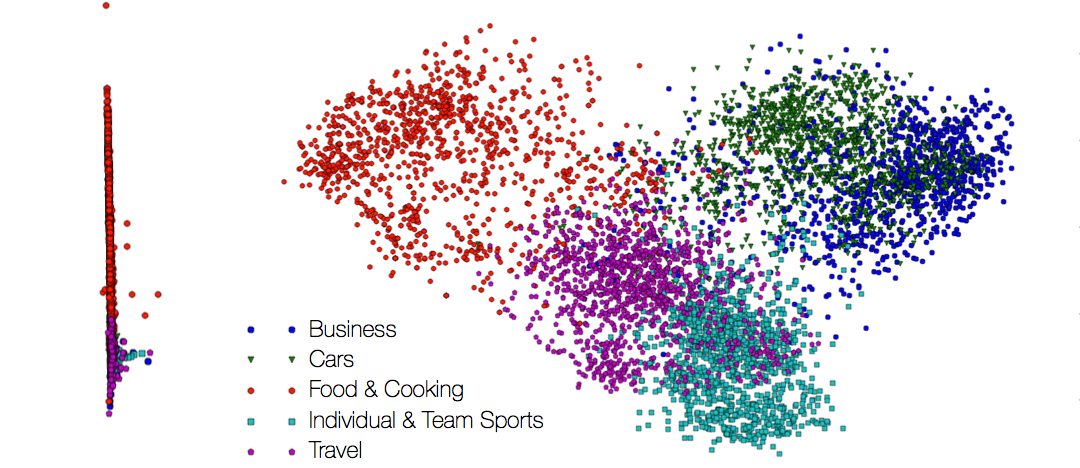}
   \caption{PCA on the 1st and 2nd principal components on the test dataset input vectors and output vectors from a 2000-500-250-125-10-DBN. \textbf{Left:} PCA on the $2000$-dimensional input. \textbf{Right:} PCA on the $10$-dimensional output.}
   \label{fig:issuu_pca}
\end{figure}

When plotting the test dataset output vectors of the 2000-500-250-125-10-DBN for the 1st and 2nd principal component, it is evident how the input data is cluttered and how the DBN manages to spread the documents in output space according to their labels (cf. Fig. \ref{fig:issuu_pca}). By performing an analysis of the output space, categories such as \textit{Business} and \textit{Cars} are in close proximity to each other and far from a category like \textit{Food \& Cooking}.

Exploratory data analysis on the Issuu Corpus show how the 2000-500-250-125-10-DBN maps documents into output space\footnote{Due to copyright issues and the terms of services/privacy policy at Issuu the results are not visualized in this article.}. We have chosen random query documents from different categories and retrieved 10 documents within the nearest proximity. When we query a car publication about an \textit{SUV}, the 10 documents retrieved from output space are about cars. They are all publications promoting a new car, published by the car manufacturer. 7 out of the 10 related publications concern the same type of car. When comparing a query in output space with the same query in the high-dimensional input space, we see that the similar documents are more accurate in output space from a human perception.


\section{Conclusion}

On the Wikipedia and Issuu corporas we have shown how the DBN is superior compared to the proposed LDA models. The DBN manages to find a better internal representation of the documents in an output space of lower dimensionality. The low dimensionality of the output space results in fast retrieval of similar documents. A binary output vector of a larger dimensionality performs almost as good as a real valued output vector of equivalent dimensionality. Finding similar documents from binary latent representations is even faster.
 

\bibliography{example_paper}
\bibliographystyle{icml2014}

\end{document}